# Lexical Resources for Hindi Marathi MT

**Sreelekha. S, Pushpak Bhattacharyya**

IIT Bombay

E-mail: {sreelekha, pb}@cse.iitb.ac.in

**Abstract**

In this paper we describe some ways to utilize various lexical resources to improve the quality of statistical machine translation system. We have augmented the training corpus with various lexical resources such as IndoWordnet semantic relation set, function words, kridanta pairs and verb phrases etc. Our research on the usage of lexical resources mainly focused on two ways such as augmenting parallel corpus with more vocabulary and augmenting with various word forms. We have described case studies, evaluations and detailed error analysis for both Marathi to Hindi and Hindi to Marathi machine translation systems. From the evaluations we observed that, there is an incremental growth in the quality of machine translation as the usage of various lexical resources increases. Moreover usage of various lexical resources helps to improve the coverage and quality of machine translation where limited parallel corpus is available.

## 1. Introduction

Machine Translation (MT) is the process of translating text or speech from one natural language to another with the help of machines. There are many ongoing attempts to develop MT systems for various Indian languages using rule-based as well as statistical-based approaches. Since India is rich in linguistic divergence there are many morphologically rich languages quite different from English as well as from each other, there is a great need for machine translation between them (Nair et.al, 2012). It has 22 scheduled languages, which are written in 10 different scripts. This paper discusses various approaches used in Indian language to Indian language MT systems especially in Marathi to Hindi Statistical MT system and vice versa to improve the quality of machine translation.

For example, consider the translated Hindi output from Marathi-Hindi SMT system for the Marathi sentence,

*Marathi -* ह्या कारणास्तव तो नाराज होता.
     *{hya karnaastav to naraz hota}*

*Hindi-* यह फलस्वरूप वह नाराज हो ।
     {yah phalswaroop vah naraz ho}

Here the Marathi word "कारणास्तव"{*karnaastav*}*is wrongly mapped to Hindi word* "फलस्वरूप"{*phalswaroop*} *and the verb* होता{*hota*} *is* wrongly mapped to हो गया {*ho gaya*}.

In order to learn various inflected and verb forms, lexical resources can play a major role. The detailed analysis about various linguistic phenomena and how the lexical resources can be used for improving the translation quality is explained in the following sections.

In the case of Marathi, a single noun can have more than 200 forms that are either adjectives or adverbs. Similarly, a verb may exhibit over 450 forms. Also the language is covering about 10,000 nouns and over 1,900 verbs. Moreover 175 postpositions can be attached to nominal and verbal entities. Some postpositions can occur in compound forms with most of the other postpositions. In addition, there are many kinds of derivable words such as causative verbs like '*karavane*', i.e. 'to make (someone) to do (something)', which is derivable from root '*karane*' i.e. 'to do', and abstract nouns like '*gharpan*' i.e. 'homeliness', which is derivable from '*ghar*' i.e. 'home'[Veena et.al., 2005].

Major difficulties in Machine Translation are handling the structural difference between the two languages and handling the ambiguities.

### 1.1 Challenge of Ambiguity

There are two types of ambiguity: structural ambiguity and lexical ambiguity.

#### 1.1.1. Lexical Ambiguity

Words and phrases in one language often have multiple meaning in another language.
For example, in the sentence,

*Marathi-* **आले होते म्हणून काम झाले**
     {*aale hote mehnaan kaam jchale*}
*Hindi-* अदरक था इसलिए काम हुआ /
     {*aadark tha isliye kaam hua*}
*English-* Ginger was there so the work has done
     *or*
*Hindi-* आए थें इसलिए काम हुआ /
     {*maenne photo nikala*}
*English-* came thatsy work was done

Here in the above sentence "*आले*"{aale}, has ambiguity in meaning. It is not clear that whether the word "*आले*"{aale}, is used in Hindi as the "ginger" ("*अदरक*"{ aadark } sense or the "*आए*"{aayen}, sense. This kind of ambiguities will be clear from the context.

**1.1.2. Structural Ambiguity**

In this case due to the structural order there will be multiple meanings. For example,

Marathi - तिथे लठ्ठ मुली आणि मुले होती.
      {tithe latth muli aani mulen hoti}
      *{There were fat girls and boys there}*

Here from the words "लठ्ठ मुली आणि मुले" {lattha muli aani mulen } it is clear that fat girls but it is not clear that boys are fat, since in Marathi to represent tall girls and boys only one word "लठ्ठ" {lattha} {tall} is being used. It can have two interpretations in Hindi and English according to its structure.

Hindi - वहाँ मोटी लड़कियां और लडकें थे ।
    *{vahan moti ladkiyam our ladkem the }*
    *{There were fat girls and boys there}*
    or
Hindi - वहाँ मोटी लड़कियां और मोटे लडकें थे ।
    *{vahan moti ladkiyam our mote ladkem the}*
    *{There were fat girls and fat boys there}*

To generate appropriate Machine Translations by handling this kind of structural ambiguity is one of the big problems in Machine Translation.

**1.2 Structural Differences**

In the case of Marathi – Hindi machine translation both languages follow the same structural ordering in sentences, such as Subject- Object-Verb (SOV). Even though there is ordering similarity, there are morphological and stylistic differences which have to be considered during translation. Marathi is morphologically more complex than Hindi, wherein there are a lot of post-modifiers in the former as compared to the later (Dabre *et al,* 2012, Bhosale, 2011).

For example, the word form "हस्तिनापुरच्या"*{hustinapoorchya}* *{of / about the hustinapuri}* is derived by attaching "च्या"*{chyaa}* as a suffix to the noun "हस्तिनापुर" *{hustinapur}{a place in India}* by undergoing an inflectional process. Marathi exhibits agglutination of suffixes which is not present in Hindi and therefore these suffixes has equivalents in the form of post positions. For the above example, the Hindi equivalent of the suffix "च्या" *{chyaa}* is the post position "के"*{ke}* which is separated from the noun "हस्तिनापुर"*{ hustinapur}.* Hence the translation of "हस्तिनापुरच्या"*{ hustinapoorchya }* will be "हस्तिनापुर के" *{ hustinapur ke}.*

Consider an example for word ordering ,

Hindi- गढ़-मुक्तेश्वर हिंदुओं का पावन तीर्थ है ।
    *{gad-mukteshwar hinduom ka paavan teerth hai }*
    (S)    (O)    (V)
Marathi- *गढमुक्तेश्वर हिंदुंचे पवित्र तीर्थ आहे .*
    *{gadmukteshwar hindoonche pavitr teerth aahe}*
    (S)    (O)    (V)
*English-Gadmuktheshwar is the holy place for Hindu.*
    (S)    (V)    (O)

Since the word order is same for both the languages it is an advantage for statistical machine translation system during alignment. And it will improve the quality of translation.

**1.2.1 Participial Constructions**

Constructions in Hindi having Participials in Marathi
Example: Hindi:

जो लड़का गा रहा था वह चला गया
 jo ladkaa gaa rahaa thaa wah chalaa gayaa
 rel. boy sing stay+perf.+cont. be+past walk go+perf.
 The boy who was singing, has left.

Marathi (Direct Translations):
   जो मुलगा गात होता तो निघून गेला
   jo mulgaa gaat hotaa to nighoon gelaa
   rel. boy sing+imperf. be+past leave+CP go+perf.
   The boy who was singing, has left.

Participial Constructions in Marathi(Actual Translations):
   गाणारा मुलगा निघून गेला
   gaaNaaraa mulgaa nighoon gelaa
   sing+part. boy leave+CP go+perf.
   The boy who was singing left

Note:-
   Deletion/dropping of subordinate clause is common in Marathi as compared to Hindi.

**1.3 Vocabulary Differences**

Languages differ in the way they lexically divide the conceptual space and sometimes no direct equivalents can be found for a particular word or phrase of one language in another.
Consider the sentence,
    काल मंगळागौरीची पूजा होती.
    *{kaal mangalgowreechi pooja hoti}*
{Yesterday the pooja which happens in the month of sravan for married women got completed}

*Here* "मंगळागौरीची"{*mangalagowrichi*}{the pooja which happens in the month of sraven for married women} as a verb has no equivalent in Hindi, and this sentence has to be translated as,
"कल सुहागन की श्रावण मास में संपन्न होनेवाली पूजा थी|
*{"Kal sahagan ki sravan maas mem sampannu honewali pooja thi."}*
*{* Yesterday the pooja which happens in the month of sravan for married women got completed *}*
   To determine translations of such language specific concepts pose additional challenges in machine translation.
.

# 2. Difficulties in SMT approach

As described in section 1 there are many structural differences between languages. The statistical approach tries to generate translations based on the knowledge and statistical models extracted from parallel aligned bilingual text corpora. Statistical models take the assumption that every word in the target language is a translation of the source language words with some probability (Brown et al., 1993). The words which have the highest probability will give the best translation. Consistent patterns of divergence between the languages (Dorr et al., 1994, Dave et al., 2002, Ramananthan et al. , 2011) when translating from one language to another, handling reordering divergence are one of the fundamental problems in MT (Kunchukuttan and Bhattacharyya 2012, Och and Ney, 2001, Koehn, 2007).

In the case of Marathi, Hindi even though both the language follows same word order there are structural difference between the language and in the generation of word forms due to the morphological complexity as described in section 1. In order to overcome this difficulty and make the machine to learn different morphological word forms lexical resources can play a major role. Different word forms such as verb phrases, morphological forms etc can be used. Also the SMT system lacks in vocabulary due to the small amount of parallel corpus. Comparative performance studies conducted by Och and Ney have shown the significance of adding dictionary words into corpus and the improvement in the translation quality in their paper (Och, Ney 2003). In order to increase the coverage of vocabulary we have used IndoWordnet. IndoWordnet(Bhattacharyya, 2010) is a lexical database for various Indian languages, in which Hindi wordnet is the root and all other Indian language wordnets are linked through the expansion approach. Words and its concepts are stored in a structure called the Lexical Matrix, where rows represent *word meanings* and columns represents the *forms*. The extraction of bilingual mapped words and its usage in machine translation is described in the experimental section 3. The comparative performance analysis with phrase based model and the phrase based model after augmenting various lexical resources is described in section 4.
.

## 3. Experimental Discussion

We now describe the development of our Marathi- Hindi and Hindi- Marathi SMT system[1] (Sreelekha, Dabre and Bhattacharyya 2013), the experiments performed and the comparisons of results, in the form of an error analysis. For the purpose of constructing with statistical models we use Moses and Giza++[2]. We have conducted various experiments to improve the quality of machine translation by utilizing various lexical resources.

Our experiments are listed below:

---

[1] http://www.cfilt.iitb.ac.in/SMTSystem
[2] http://www.statmt.org/

1. SMT system with an uncleaned corpus
2. SMT system with a cleaned corpus
3. Corpus with IndoWordnet extracted words
4. Corpus with Function words and Kridantha pairs
5. Corpus with verb phrases

### 3.1.1 SMT system with an uncleaned corpus

We have used 50000 corpus of Tourism and Health provided by ILCI consortium. There were many misalignments, wrong and missing translations in the corpus. It affected the translation and the quality was not good.

For example, sentence from the corpus where the translation is wrong,

**Hindi** :
मणिपुर के दूरस्थ उत्तर-पूर्वी राज्य में पोलो खेल का अस्तित्व कायम रहने के लिए, संसार आधुनिक पोलो के जन्म का ऋणी है, क्योंकि आज यह पूरी दुनिया में खेला जा रहा है ।
*{manipur ke doorsth uththa-poorv raajy mem polo khel ka astivthv kaayam rahne keliye, samsar aadhunik polo ke janmu ka shruni hae, kyonki aaj yah poori duniya mem khela ja raha hae |}*
**English** : *{To retain the existence of the game of Polo in the far north eastern state of Manipur, the world is indebted to the birth of modern polo as today it is being played all over the world.}*

**Equivalent Marathi Translation** (wrong)
पोलो आज जगभर खेळला जातो त्याचे श्रेय तो ईशान्य भारतातील मणिपूरच्या दुर्गम डोंगररांगांमध्ये जिवंत राहिला याकडे जाते.
{*polo aaj jagbhar khelela jato tyache shrey to yieshany bharatateel manipoorchya durgm dongarangamadye jivant raahila yakade jaate*}
**English:** *Today Polo is being played all over the world and it's credit goes to retaining the existence of the game of Polo in the remote hills of Manipur of north eastern India.*

The comparative performance results of cleaned corpus over uncleaned corpus were shown in the Table 4, 5, 6 and 7. From the error analysis we came to a conclusion that in order to improve the translation quality we need to provide a well cleaned parallel corpus for training.

### 3.1.2 SMT system with cleaned corpus

We have corrected the misalignments between parallel sentences which improves the learning of word to word alignments. Sometimes we had to correct source sentences even.

For example, for the above sentence
Hindi :
मणिपुर के दूरस्थ उत्तर-पूर्वी राज्य में पोलो खेल का अस्तित्व कायमं रहने के लिए, संसार आधुनिक पोलो के जन्म का ऋणी है, क्योंकि आज यह पूरी दुनिया में खेला जा रहा है ।
{*manipur ke doorsth uththa-poorv raajy maem polo khel ka astivthv kaayam rahne keliye, samsar aadhunik polo ke janmu ka shruni hae, kyomki aaj yah poori duniya mem khela ja raha hae* }

**Correct Marathi Translation**

मणिपूरच्या दूरस्थ ईशान्य राज्यांमध्ये पोलो खेळाचे अस्तित्व टिकून ठेवण्यासाठी, संसार आधुनिक पोलोच्या जन्माचे ऋणी आहे, कारण की आज हा संपूर्ण जगभर खेळला जात आहे.
*{manipoorchyaa durasth eeshanya rajaymadye polo khelache asthvithv tikoon thevnyasati, samsar aadhunik polochya jenmache shruni aahe, kaaran ki aaj ha sampoorn jagphar khetla jaat aahe.}*
**English** : *{To retain the existence of the game of Polo in the far north eastern state of Manipur, the world is indebted to the birth of modern polo as today it is being played all over the world.}*

We have removed the stylistic constructions from the parallel corpus which prevent the learning of grammatical structures. Removed the unwanted characters and wrong translations and corrected the missing translations and phrases. The resultant machine translation system's quality improved more than 40 %. The statistics of corpus used are shown in Table 1 and the results are shown in the Tables 4, 5, 6 and 7. During error analysis we came to know that the machine lacks in sufficient amount of vocabulary and hence we investigated on the usage of IndoWordnet to improve the quality of machine translation.

### 3.1.3 Corpus with IndoWordnet extracted words

We have extracted a total of 437832 parallel Marathi-Hindi words using bilingual mapping according to its semantic and lexical relations as described in section 2. We have used an algorithm to extract the bilingual words from IndoWordnet. Bilingual mappings are generated using the concept-based approach across words and synsets (Mohanty et.al., 2008). For a single word considered it's all synset word mappings and generated that many entries of parallel words.

For example, the word अंतहीन{*antaheen*} has the following equivalent synset words in IndoWordnet.

अंतहीन: अनंत असमाप्य अंतहीन अनन्त अन्तहीन अनवसान

*{antaheen: anantu asamapya antaheen anant antaheen anavasaan}*
*{endless: endless not-ending endless infinite endless not-ending }*

We augmented the extracted parallel words into the training corpus. It helped in improving the translation quality to a great extent. The statistics of Wordnet synsets used are shown in Table 2 and the results are shown in the Tables 4, 5, 6 and 7. During the error analysis we observed that even though the machine translation system is able to give considerably good quality translation, it lacks in handling case markers and inflected forms. One of the advantages is that it helped the machine to handle word sense disambiguation well, since the synsets covers all common forms of a word.

| Sl.No | Corpus Source | Training Corpus [Manually cleaned and aligned] | Corpus Size [Sentences] |
|---|---|---|---|
| 1 | ILCI | Tourism | 25000 |
| 2 | ILCI | Health | 25000 |
| 3 | DIT | Tourism | 20000 |
| 4 | DIT | Health | 20000 |
| | | Total | 90000 |

Table 1: Statistics of Training Corpus

| Sl.No | Lexical Resource Source | Lexical Resources in Corpus | Lexical Resource Size [Words] |
|---|---|---|---|
| 1 | CFILT, IIT Bombay | IndoWordnet Synset words | 437832 |
| 2 | CFILT IIT Bombay | Function Word, Kridanata Pairs | 5000 |
| 3 | CFILT IIT Bombay | Verb Phrases | 4471 |
| | | Total | 447303 |

Table 2: Statistics of Lexical Resources Used

| Sl. No | Corpus Source | Tuning corpus [Manually cleaned and aligned] | Corpus Size [Sentences] |
|---|---|---|---|
| 1 | EILMT | Tourism | 100 |
| | | Total | 100 |

Table 3: Statistics of Tuning Corpus

| Sl. No | Corpus Source | Testing corpus [Manually cleaned and aligned] | Corpus Size [Sentences] |
|---|---|---|---|
| 1 | EILMT | Tourism | 100 |
| | | Total | 100 |

Table 4: Statistics of Testing Corpus

### 3.1.4 Corpus with Function words and Kridantha pairs

Marathi and Hindi have 7 types of kridanta forms, its' post position, pre-position and inflected forms. We have augmented kridanta, akhyat, function words, suffix pairs etc into the training corpus. This helped in machine translation system to infer the grammatical structures and hence the quality of translation improved.

Sample Marathi-Hindi kridanta form pair is,

आयला_पाहिजे : _ना_चाहिए
{aayla pahije : na chahiye}
{need : need}

The statistics of function words used are shown in Table 2 and the results are shown in the Tables 4, 5, 6 and 7. From the error analysis we came to a conclusion that translation system is facing difficulties in handling verbal translations because of the morphological phenomena.

### 3.1.5 Corpus with verb phrases

In order to overcome the verbal translation difficulty we have extracted Marathi- Hindi parallel verbal forms and its translations which contains various phenomena. We have augmented the 4471 entries of verbal translations into the training corpus.

Sample Marathi-Hindi verb form pair is,

जरूर करवा लें : अवश्य करवून घ्या
{zaroor karva lem : avasyu karvoon ghya}
{should get it done : should get it done}

The statistics of function words used are shown in Table 2 and the results are shown in the Tables 4, 5, 6 and 7. The error analysis study has shown that the quality of the translation has been improved drastically.

## 4. Evaluation

We have tested the translation system with a corpus of 100 sentences taken from the 'EILMT tourism health' corpus as shown in Table 3. We have used various evaluation methods such as subjective evaluation, BLEU score (Papineni et al., 2002), METEOR and TER (Agarwal and Lavie 2008). But we gave importance to subjective evaluation to determine the fluency (F) and adequacy (A) of the translation, since for morphologically rich languages subjective evaluations can give more accurate results comparing to other measures. Fluency is an indicator of correct grammatical constructions present in the translated sentence whereas adequacy is an indicator of the amount of meaning being carried over from the source to the target. Marathi Hindi Bilingual experts have assigned scores between 1 and 5 for each generated translation, on the basis of how much meaning conveyed by the generated translation and its grammatical correctness. The basis of scoring is given below:

- 5: The translations are perfect.
- 4: One or two incorrect translations and mistakes.
- 3: Translations are of average quality, barely making sense.
- 2: The sentence is barely translated.
- 1: The sentence is not translated or the translation is gibberish.

$S1, S2, S3, S4$ and $S5$ are the counts of the number of sentences with scores from 1 to 5 and $N$ is the total number of sentences evaluated. The formula (Bhosale et al., 2011) used for computing the scores is:

$$A/F = 100 * \frac{(S5 + 0.8*S4 + 0.6*S3)}{N}$$

We consider only the sentences with scores above 3. We penalize the sentences with scores 4 and 3 by multiplying their count by 0.8 and 0.6 respectively so that the estimate of scores is much better. The estimate may vary from person to person as these scores are subjective, in which case an inter annotator agreement is required. We do not give these scores, since we had only one evaluator. The results of our evaluations are given in Table 4, 5, 6 and 7.

## 5. Error Analysis

We have evaluated the translated outputs of both Marathi to Hindi and Hindi to Marathi Statistical Machine Translation systems in all 5 categories as explained in Section 3. The detailed error analysis is shown in Table 8 and 9 for a sentence exhibiting a variety of linguistic phenomena and how the quality of Machine Translation system changes by augmenting various lexical resources. The results of BLEU score, METEOR and TER evaluations are displayed in Tables 6 and 7 and the results of subjective evaluations are displayed in Table 4 and 5. We have observed that the quality of the translation is improving as the corpus is getting cleaned as well as more lexical resources are used. Hence there is an incremental growth in adequacy, fluency, BLEU score, METEOR score and in TER score. The fluency of the translation increased up to 83% in the case of Marathi to Hindi and up to 85% in the case of Hindi to Marathi.

Also we have observed that the score of Hindi to Marathi translation quality is slightly higher than that of Marathi to Hindi translation. Since Marathi is morphologically richer than Hindi and Marathi have more agglutinative suffixes attached, while in Hindi it is not present, as explained in above section 1.2. Therefore these Marathi suffixes have Hindi equivalents in the form of post positions. So during alignment from Hindi to Marathi, Hindi word can align to the words with agglutination in Marathi, since it is a single word. But on the other hand while aligning form Marathi-Hindi the agglutinative word can map to only root words, there is a chance to miss out the post position mapping, since it is separate words. So it will improve the translation quality of Hindi- Marathi SMT as compared to Marathi-Hindi SMT.

For example, during Hindi-Marathi Translation, in Hindi-Marathi alignment the noun "हस्तिनापुर"

*{hustinapur}* can easily align to "हस्तिनापुरच्या" *{hustinapoorchya}*, since it is a single word. But in Marathi-Hindi alignment, "हस्तिनापुरच्या" *{hustinapoorchya}* may be map to "हस्तिनापुर" *{hustinapur}* and there is a possibility to drop down the post position "के"*{ke}* which is a separate word from the noun "हस्तिनापुर"*{hustinapur}*. These features will have impact on the translation quality and hence the inflected forms may not translate properly from Marathi to Hindi. On the other hand, Hindi to Marathi translation system will not face this difficulty because of alignments. Thus there is an improvement in quality of Hindi-Marathi MT comparing to Marathi-Hindi MT.

| Marathi-Hindi Statistical MT System | | Adequacy | Fluency |
|---|---|---|---|
| With Uncleaned Corpus | With Tuning | 20.6% | 30.8% |
| | Without Tuning | 17.8% | 24% |
| With Cleaned Corpus | With Tuning | 58.6% | 68.3% |
| | Without Tuning | 54% | 64% |
| With Wordnet | With Tuning | 72.4% | 80% |
| | Without Tuning | 69.6% | 78.2% |
| With Function Words, kridanta pairs | With Tuning | 78% | 81% |
| | Without Tuning | 75% | 80% |
| With verb Phrases | With Tuning | 83% | 88% |
| | Without Tuning | 80% | 85% |

Table 4: Results of Marathi-Hindi SMT Subjective Evaluation

| Hindi-Marathi Statistical MT System | | Adequacy | Fluency |
|---|---|---|---|
| With Uncleaned Corpus | With Tuning | 22.87% | 31.3% |
| | Without Tuning | 20.56% | 28% |
| With Cleaned Corpus | With Tuning | 59% | 72.21% |
| | Without Tuning | 53% | 65% |
| Corpus with Wordnet | With Tuning | 73% | 83% |
| | Without Tuning | 70.01% | 80.04% |
| Corpus with Function words, kridanta pairs | With Tuning | 79.36% | 87.21% |
| | Without Tuning | 76% | 85.68% |
| Corpus with verb Phrases | With Tuning | 85.01% | 89.32% |
| | Without Tuning | 82% | 86.34% |

Table 5: Results of Hindi-Marathi SMT Subjective Evaluation

| Hindi-Marathi Statistical MT System | | BLEU score | METEOR | TER |
|---|---|---|---|---|
| With Uncleaned Corpus | With Tuning | 1.96 | 0.124 | 45.06 |
| | Without Tuning | 1.26 | 0.127 | 44.55 |
| With Cleaned Corpus | With Tuning | 7.76 | 0.193 | 84.05 |
| | Without Tuning | 3.97 | 0.190 | 83.92 |
| Corpus with Wordnet | With Tuning | 11.78 | 0.225 | 82.91 |
| | Without Tuning | 9.31 | 0.217 | 84.30 |
| With Function Words, kridanta pairs | With Tuning | 12.21 | 0.274 | 83.79 |
| | Without Tuning | 9.25 | 0.214 | 85.06 |
| With verb Phrases | With Tuning | 18.15 | 0.281 | 77.94 |
| | Without Tuning | 10.26 | 0.261 | 85.48 |

Table 6 : Results of Hindi-Marathi SMT BLEU score, METEOR, NER Evaluations

| Marathi-Hindi Statistical MT System | | BLEU score | METEOR | TER |
|---|---|---|---|---|
| With Uncleaned Corpus | With Tuning | 1.86 | 0.119 | 50.32 |
| | Without Tuning | 1.10 | 0.105 | 49.52 |
| Corpus with Cleaned Corpus | With Tuning | 8.01 | 0.171 | 78.32 |
| | Without Tuning | 4.22 | 0.160 | Infinity |
| Corpus with Wordnet | With Tuning | 11.54 | 0.278 | 80.75 |
| | Without Tuning | 9.78 | 0.226 | 77.30 |
| Corpus with Function Words, kridanta pairs | With Tuning | 12.20 | 0.283 | 81.19 |
| | Without Tuning | 10.46 | 0.263 | 81.19 |
| Corpus with verb Phrases | With Tuning | 17.80 | 0.288 | 80.36 |
| | Without Tuning | 13.70 | 0.265 | 81.48 |

Table 7: Results of Marathi-Hindi SMT BLEU score, METEOR, NER Evaluations

| Sr. No. | | | Sentence | Explanation of phenomena |
|---|---|---|---|---|
| 1 | Source Sentence : Hindi | | केंद्रीय सरकारी संग्रहालय १८७६ में वेल्स के राजकुमार के भारत दर्शन के समय बनवाया गया था और 1886 में जनता के लिए खुला था। | |
| | Meaning | | In 1986 the national central museum was established during the visit of the Prince of Wales and in 1886 was opened for the public. | |
| | With Uncleaned Corpus | With Tuning | केंद्रीय सरकारी संग्रहालय १८७६ साली वेल्स गवताळ राजकुमार के भारत दर्शन वेळी घेतलेला के शेड डाला गया 1886 साली जनतेसाठी खुले. | 1. Many words are wrongly translated. 2. Function words not translated: के 3. Verb translation has tense problem: था not translated. 5. Insertion case: के शेड डाला, गवताळ 6. Conjunction not translated. |
| | | Without Tuning | केंद्रीय सरकारी संग्रहालय १८७६ मध्ये वेल्स ह्याच्या राजकुमार के भारत दर्शन वेळी घेतलेला होती 1886 मध्ये वह जनतेसाठी खुला था . | |
| | With Cleaned Corpus | With Tuning | केंद्रीय सरकारी संग्रहालय १८७६ मध्ये वेल्स करण्यासाठी राजकुमाराच्या भारतात दर्शन वेळ बनवले आणि 1886 मध्ये जनतेसाठी खुले होता. | 1. Function word: के not translated 2. In the second verb part, there is an inflection problem खुले होता 3. Insertion verb case: करण्यासाठी 4. Inflection problem: "दर्शन के समय" not translated correct as: दर्शनाच्या वेळी 5. First verb not translated correctly केले होते |
| | | Without Tuning | केंद्रीय सरकारी संग्रहालय १८७६ मध्ये वेल्स ह्याच्या राजकुमाराच्या भारत दर्शन करताना केले होते 1886 मध्ये जनतेसाठी खुला होता. | |
| | Corpus With Wordnet | With Tuning | केंद्रीय सरकारी संग्रहालय १८७६ मध्ये वेल्स येथील राजकुमाराच्या भारत दर्शन वेळ बनवले होते आणि 1886 मध्ये जनतेसाठी खुले. | "दर्शनच्या" suffix addition is missing. Verb "होते" is wrongly translated as "होती". "वेल्स येथील" is correctly translated. |
| | | Without Tuning | केंद्रीय सरकारी संग्रहालय १८७६ मध्ये वेल्स येथील राजकुमाराच्या भारत दर्शन वेळी बनवले होते आणि 1886 मध्ये जनतेसाठी खुले होती. | |
| | Corpus with Function Words, kridanta pairs | With Tuning | केंद्रीय सरकारी संग्रहालय १८७६ मध्ये वेल्स येथील राजकुमाराच्या भारत दर्शनच्या वेळी बनवले होते आणि 1886 मध्ये जनतेसाठी खुले होता. | "दर्शनच्या" suffix addition is missing. "वेल्स येथील" got translated correctly. Verb "होता" didn't get translated. |
| | | Without Tuning | केंद्रीय सरकारी संग्रहालय १८७६ मध्ये वेल्स येथील राजकुमाराच्या भारत दर्शनच्या वेळी बनवले होते आणि 1886 मध्ये जनतेसाठी खुले होते | |
| | Corpus with verb Phrases | With Tuning | केंद्रीय सरकारी संग्रहालय १८७६मध्ये प्रिंस ऑफ वेल्सच्या राजकुमार याच्या भारतभेटीच्या वेळी उभारण्यात आले होते व १८८६ मध्ये जनतेसाठी खुले होते . | "राजकुमार याच्या" is translated correctly. Verb "होते" didn't get translated. |
| | | Without Tuning | केंद्रीय सरकारी संग्रहालय १८७६ मध्ये वेल्स येथील राजकुमार याच्या भारत दर्शनच्या वेळी बनवले होते आणि 1886 मध्ये जनतेसाठी खुले होते . | |

**Table 8: Hindi- Marathi SMT Error Analysis**

| Sr. No. | Sentence | | | Explanation of phenomena |
|---|---|---|---|---|
| 1 | Source Sentence : Marathi | | एखादा भरतपूरमध्ये एका ठिकाणापासून दुसरीकडे जाण्यासाठी टॅक्सी, सायकलरिक्षा व ऑटोरिक्षा यासारखे वाहतुकीचे अनेक पर्याय घेऊ शकतो . | |
| | Meaning | | In Bharatpur for going from one place to another can bring taxi, cycle rickshaw or auto rickshaw. | |
| | With Uncleaned Corpus | With Tuning | कोई भरतपूरमध्ये एक स्थान दूसरी ओर जाने टॅक्सी, सिसोदिया ऑटोरिक्षा यासारखे ह्यांच्या कार्य कई विकल्प सकता है । | Here the function words "के लिए", "से", verb " घेऊ", Conjunction " व", Insertion cases: कार्य, ह्यांच्या etc are not translated. Also Modal Auxiliary verb "शकतो" is translated as "सकते" instead of "सकता है ।" Post-positions not translated for words मध्ये, पासून, कडे, साठी. Wrong translated word: सिसोदिया. Not translated words: यासारखे, वाहतुकीचे |
| | | WithoutTuning | कोई भरतपूरमध्ये एक स्थान दूसरी ओर जाने टॅक्सी, सिसोदिया ऑटोरिक्षा यासारखे ह्यांच्या कार्य यासारखे वाहतुकीचे कई विकल्प सकते | |
| | With Cleaned Corpus | With Tuning | कोई भरतपुर में एक सदर से दूसरी ओर के लिए टॅक्सी, सायकलरिक्षा और ऑटोरिक्षा जैसी यातायात पानी से कर सकता है । | Here the पर्याय घेऊ is wrongly translated as Insertion case "पानी से कर". The kridant "जाने" is missing in जाण्यासाठी translation, only post-position साठी translated. Missing word: जाण्या. Suffix चे not translated (वाहतुकीचे). Wrongly translated words-ठिकाण – सदर, अनेक पर्याय. Verb wrongly translated: घेऊ– कर |
| | | Without Tuning | कोई भरतपुर में एक स्थान से दूसरी ओर के लिए टॅक्सी, सायकलरिक्षा और ऑटोरिक्षा जैसी यातायात पानी से कर सकता है । | |
| | Corpus With Wordnet | With Tuning | कोई भरतपुर में एक स्थान दूसरी और टैक्सी, रिक्शा व आटो_रिक्शा जैसी यातायात कई विकल्प ले | Krudant form "जाण्यासाठी" not transferred. Modal Auxiliary verb "शकतो" is translated as plural form "सकते" instead of the kridanta "सकता है". Function word "से", "केलिए", "के", are missing. Suffix "चे" not translated from वाहतुकीचे. |
| | | Without Tuning | कोई भरतपुर में एक स्थान से दूसरी ओर जाने के लिए टैक्सी , रिक्शा व ऑटो जैसी का यातायात कई विकल्प ले सकता है । | Here the translation is perfect but the function word "जैसी का" is wrongly instead of "जैसे के". Suffix चे not translated from वाहतुकीचे. |
| | Corpus with Function Words, kridanta pairs | With Tuning | कोई भरतपुर में एक स्थान से कहीं_और जाने के लिए टॅक्सी, रिक्शा व आटो_रिक्शा जैसी यातायात के कई विकल्प ले सकता है । | The translation is good except the word "दूसरी" is misplaced as "कहीं" due to lexical choice. |
| | | Without Tuning | कोई भरतपुर में एक स्थान से दूसरी ओर जाने के लिए टॅक्सी, रिक्शा व आटो_रिक्शा जैसे यातायात के कई विकल्प ले सकता है । | |
| | Corpus with verb Phrases | With Tuning | कोई भरतपुर में एक स्थान से कहीं_और जाने के लिए टॅक्सी, रिक्शा व आटो_रिक्शा जैसी यातायात के कई विकल्प ले सकता है । | The translation is good except the word "दूसरी" is misplaced as "कहीं" due to lexical choice. |
| | | Without Tuning | कोई भरतपुर में एक स्थान से दूसरी ओर जाने के लिए टॅक्सी, रिक्शा व आटो_रिक्शा जैसी यातायात के कई विकल्प ले सकता है । | |

**Table 9: Marathi – Hindi SMT Error Analysis**

## 6. Conclusion

In this paper we have mainly focused on the usage of various lexical resources for improving the quality of Machine Translation. We have discussed the comparative performance of Statistical Machine Translation with various lexical resources for both Marathi – Hindi and Hindi-Marathi. As discussed in the experimental section, SMT, although lacks the ability to handle rich morphology it can overcome by using various lexical resources, which will help the machine to improve the translation quality.

In our experiments we have used various measures to evaluate such as BLEU Score, METEOR, TER and Fluency and Adequacy using subjective evaluation. We can see that there is an incremental growth in both Marathi- Hindi and Hindi Marathi systems in terms of BLEU-Score, METEOR, and TER evaluations. Also our subjective evaluation results shows promising scores in terms of fluency and adequacy. This leads to the importance of utilizing various lexical resources for an efficient Machine Translation system. Thus we can come to a conclusion that various lexical resources can play an important role in providing good machine translation system for morphologically rich languages.

Our future work will be focused on investigating more lexical resources for improving the quality of Statistical Machine Translation systems and there by develop an accurate MT system for both Marathi- Hindi and Hindi-Marathi Machine Translation.

## 7. Acknowledgements

We thank Almighty and truth for this work.